\documentclass[conference]{IEEEtran}
\IEEEoverridecommandlockouts
\usepackage{amsmath,amssymb,amsfonts}
\DeclareMathOperator{\logit}{logit}
\usepackage{algorithmic}
\usepackage{graphicx}
\usepackage{textcomp}
\usepackage{xcolor}
\usepackage{subcaption}
\usepackage{mathtools}
\usepackage[numbers]{natbib}
\usepackage{url}
\usepackage{hyperref}
\def\BibTeX{{\rm B\kern-.05em{\sc i\kern-.025em b}\kern-.08em
    T\kern-.1667em\lower.7ex\hbox{E}\kern-.125emX}}
\begin{document}

\title{ Causal-RetiGraph: Cross-Cohort Retinal Support and Same-Subject Pathway Analysis for Diabetic Retinopathy\\
}
\author{
\IEEEauthorblockN{Inam Ullah$^{*}$}
\IEEEauthorblockA{
University of Southampton\\
Southampton, United Kingdom\\
i1n23@soton.ac.uk
}
\and
\IEEEauthorblockN{Imran Razzak}
\IEEEauthorblockA{
Mohamed bin Zayed University of Artificial Intelligence\\
United Arab Emirates\\
imran.razzak@mbzuai.ac.ae
}
\and
\IEEEauthorblockN{Shoaib Jameel}
\IEEEauthorblockA{
University of Southampton\\
Southampton,  United Kingdom\\
M.S.Jameel@southampton.ac.uk
}
}

\maketitle
\begin{center}
{\footnotesize $^{*}$Corresponding author: i1n23@soton.ac.uk}
\end{center}
\begin{abstract}
Diabetic retinopathy (DR) is a local retinal lesion process and a visible manifestation of systemic microvascular injury. Modern retinal AI can grade images accurately, but often leaves unanswered how local lesion evidence, retinal vascular structure, and systemic disease pathways are connected. This paper introduces \emph{Causal-RetiGraph}, a compact biomedical informatics framework that links retinal graph phenotypes with NHANES-anchored pathway modelling. The retinal-image fold constructs an interpretable $X1234$ phenotype from vessel maps, lesion evidence, image embeddings, and AutoMorph biomarkers through spatial $X_{12}$ and Jacobian $X_{34}$ branches. The NHANES fold models systemic exposures, covariates, a same-subject retinal mediator family $R^*$, and downstream outcome families. $X1234$ is used for retinal support and pathway prioritisation, while $R^*$ is used for participant-level pathway summaries. On the retinal fold, $X1234$ achieves 0.9055 binary DR accuracy and 0.9711 AUROC, with graded DR QWK of 0.8312. The results show that lesion and biomarker streams improve contextual retinal representation under scarce and imbalanced
data. In NHANES, HbA1c, urine albumin, pulse pressure, fasting glucose, and systolic blood pressure are the strongest binary DR anchors. Participant-level pathway analysis identifies glycaemic--renal and glycaemic--haemodynamic pathways as the clearest mediator-style signals. These results suggest that retinal graph phenotypes can help prioritise systemic pathways in DR while preserving the distinction between image-derived support and same-subject mediation.

\end{abstract}


\section{Introduction}

Diabetes is a systemic metabolic disorder in which impaired insulin production or insulin action leads to chronic hyperglycaemia and progressive vascular injury. Over time, persistent glycaemic burden interacts with blood-pressure stress, dyslipidaemia, inflammation, renal dysfunction and disease duration, producing both macrovascular and microvascular complications \cite{antar2023diabetes, giannakogeorgou2026diabetes}. Diabetic retinopathy, diabetic nephropathy and diabetic neuropathy are among the major microvascular complications of diabetes, and they share several upstream mechanisms related to endothelial dysfunction, capillary damage and metabolic stress \cite{WHO2024Diabetes,Lu2023Vascular, Zakir2023CardiovascularDiabetes,Kulkarni2024ClinicalRelationship}. DR is therefore not only an isolated ocular disease; it is a local retinal expression of systemic diabetic microvascular burden \cite{wei2022pathophysiological}.

The retina is particularly important because it provides a non-invasive view of the human microcirculation. Unlike many vascular beds, the retinal circulation can be repeatedly imaged using colour fundus photography, allowing vessel structure, lesion burden and microvascular remodelling to be measured at scale. This has led to growing interest in retinal biomarkers and oculomics, where retinal imaging is used to study ocular, cardiovascular, renal and systemic health \cite{zhu2025oculomics}. Recent retinal biomarker roadmaps emphasise that retinal imaging may support cardiovascular and systemic disease assessment when image-derived biomarkers are standardised and linked to interpretable physiology \cite{Chew2025}. Reviews of DR imaging similarly highlight the importance of vascular calibre, vessel density, tortuosity, fractal dimension and multimodal retinal measurements for early DR characterisation and risk assessment \cite{Zhang2024}.

Automated DR screening systems have shown that fundus photographs contain sufficient information for detecting referable disease \cite{Gulshan2016}. However, the development of reliable medical image models remains constrained by data scarcity, class imbalance, inconsistent annotation quality and dataset shift. These problems are especially important in DR because severe and proliferative cases are less frequent than no-DR or mild cases, while lesion-level expert annotations are much harder to obtain than image-level labels. Recent reviews of DR deep learning and data-centric retinal AI highlight limited high-quality datasets, inconsistent labels, restricted lesion annotation and imbalance as persistent barriers to clinically reliable systems \cite{Nadeem2022DRReview,Dey2026DataCentricDR,Kwon2022DataScarcity}. Previous work on data imbalance and diversity in DR modelling has also shown that model performance can be affected by dataset composition and class distribution \cite{Inamullah2024}. These limitations motivate representations that are not only predictive, but also robust, interpretable and clinically grounded.
\begin{figure*}[t]
\centering
\includegraphics[width=\textwidth]{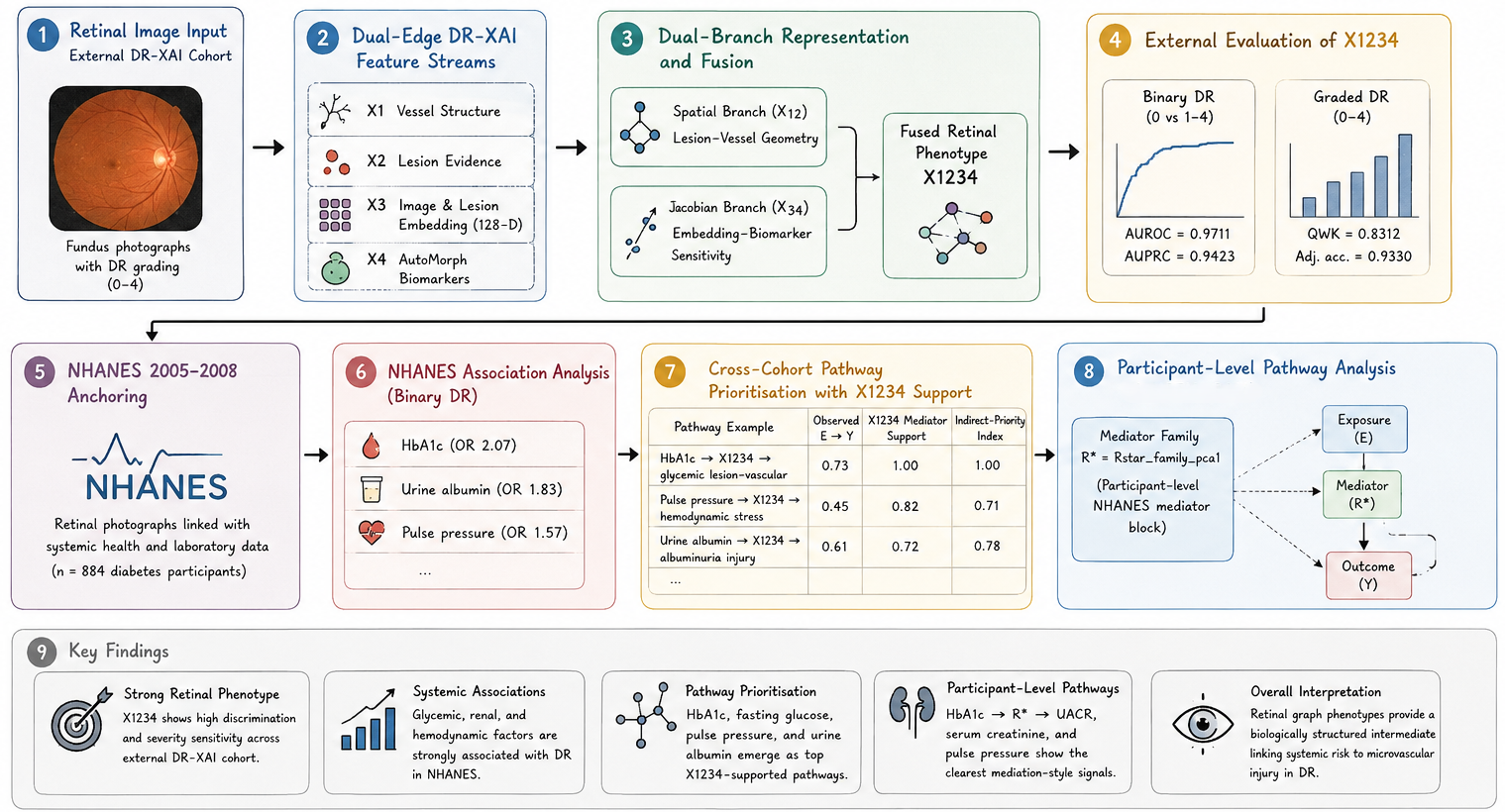}
\caption{Causal-RetiGraph workflow. Steps 1--4 construct and evaluate the image-derived retinal
    graph phenotype \(X1234\) from a DR retinal cohort. Steps 5--8 use NHANES 2005--2008 as an
    independent systemic anchoring source for exposure--DR association analysis, cross-cohort
    pathway prioritisation, and participant-level \(R^*\)-based pathway summaries. Step 9
    summarises the main framework outputs.}
\label{fig:workflow}
\end{figure*}
A second gap is interpretability. A DR grade is supported by visible lesion types, lesion burden, lesion location and vascular morphology. Microaneurysms, haemorrhages, hard exudates, cotton-wool spots and neovascularisation provide local disease evidence \cite{djoumessi2025inherently}, while retinal vessel calibre, CRAE, CRVE, AVR, vessel density, tortuosity and fractal structure describe the vascular bed~\cite{wei2022pathophysiological}. Saliency methods such as Grad-CAM can highlight class-discriminative image regions \cite{selvaraju2017}, and image-mining approaches have shown that weak heatmaps can reveal disease-relevant retinal patterns \cite{Quellec2016DeepImageMining}. However, heatmaps alone do not quantify lesion evidence \cite{ghassemi2021false}, whether it is macular, vessel-adjacent, artery-related, vein-related or biomarker-sensitive. Conversely, tabular oculomics can associate retinal vascular measurements with disease severity, but it loses the spatial arrangement of lesions and vessels. Automated tools such as AutoMorph now allow retinal vascular morphology to be quantified at scale \cite{Zhou2022AutoMorph}, and lesion-aware representation learning can improve disease-sensitive embeddings \cite{huang2021}. What remains missing is a framework that explicitly links lesion evidence, spatial vessel context, image embeddings and retinal biomarkers.

A third gap is systemic pathway interpretation. Epidemiological datasets such as NHANES \cite{cdc_nhanes} can identify relationships between systemic exposures and DR, including glycaemic, renal, haemodynamic, lipid and inflammatory factors. However, standard exposure--outcome models rarely include an interpretable retinal intermediate that connects local retinal evidence with systemic risk. This makes it difficult to ask whether retinal features behave only as passive correlates, as downstream biomarkers, or as mediator-like signals within systemic diabetic microvascular pathways. A biomedical informatics framework for DR should therefore connect three levels: retinal image evidence, retinal vascular biomarkers and participant-level systemic variables.

In light of the above, this paper presents \emph{Causal-RetiGraph}, a two-fold framework for retinal-systemic pathway analysis in DR. The first fold constructs an image-derived retinal support phenotype, denoted \(X1234\), from four evidence streams: vessel structure \((X_1)\), lesion evidence \((X_2)\), image/lesion embeddings \((X_3)\), and AutoMorph biomarkers \((X_4)\). The spatial \(X_{12}\) branch captures lesion--vessel geometry, while the Jacobian \(X_{34}\) branch captures embedding--biomarker sensitivity. These branches are fused into an interpretable retinal graph phenotype for DR grading, severity-trend analysis and lesion--biomarker hypothesis generation.

The second fold anchors systemic analysis in NHANES 2005--2008 diabetic participants. In this fold, systemic exposures \(E\), adjustment blocks \(Z/G\), retinal mediator-family variables \(R^*\), and outcome variables \(Y\) are modelled at the participant level. The key design rule is that \(X1234\) and \(R^*\) are related but not identical. \(X1234\) is an image-derived retinal support phenotype used for cross-source pathway prioritisation, whereas \(R^*\) is the same-subject NHANES retinal mediator family used for participant-level pathway summaries. This separation avoids the overclaim that an externally derived image phenotype has been measured on the same NHANES participants, while still allowing retinal image evidence to guide systemic pathway discovery.

The proposed framework operationalises the idea that the eye can serve as a window to systemic health. Rather than treating the retina only as an image-classification input or only as an epidemiological outcome, Causal-RetiGraph treats retinal evidence as a structured biomedical layer for pathway prioritisation and hypothesis generation. This may support pathway prioritisation, reveal lesion--vascular coupling, and motivate future same-subject tests of whether retinal graph phenotypes behave as correlates, mediator-like variables or modifiers of systemic diabetic risk.

The contributions are fourfold. First, we construct a retinal graph phenotype \(X1234\) that combines spatial lesion--vessel evidence and Jacobian embedding--biomarker sensitivity. Second, we build an NHANES binary DR anchor to identify glycaemic, renal and haemodynamic systemic factors associated with DR. Third, we introduce a cross-source prioritisation strategy that combines observed NHANES exposure--DR effects with retinal \(X1234\) support. Fourth, we report participant-level pathway summaries through NHANES \(R^*\), identifying glycaemic--renal and glycaemic--haemodynamic axes as dominant DR-related pathway families.

\section{Related Work}
\subsection{DR AI and explainability}
Deep learning systems have achieved strong performance for DR screening from retinal fundus photographs \cite{gargeya2017}. Their usefulness in practice depends not only on classification accuracy but also on whether the output can be related to clinically meaningful evidence. Grad-CAM and related class activation methods are valuable for weak localisation, but they should not be interpreted as expert lesion segmentation \cite{selvaraju2017}. Lesion-aware representation learning offers a complementary direction by shaping embeddings around disease-relevant visual structures \cite{huang2021}. Causal-RetiGraph uses these ideas to build a retinal phenotype that can be linked to pathway-level analysis rather than only to class prediction. Unlike standard DR screening models, Causal-RetiGraph uses image-derived evidence to support pathway-level analysis rather than only disease classification.

\subsection{Retinal biomarkers and oculomics}
Retinal imaging biomarkers provide a quantitative link between the eye and systemic vascular biology. AutoMorph extracts artery, vein and vessel morphology features at scale, including calibre, density, tortuosity, fractal dimension, CRAE, CRVE and AVR. Biomarker roadmaps argue that retinal measurements can contribute to cardiovascular and systemic disease assessment when they are standardised and physiologically interpretable \cite{zhu2025oculomics}. Structural and functional DR imaging reviews further motivate early biomarker discovery for diabetic eye disease \cite{wei2022pathophysiological}. These developments motivate retinal phenotypes that combine images, lesions and vessel biomarkers. Unlike purely tabular oculomics analyses, Causal-RetiGraph retains lesion--vessel spatial context through
the \(X_{12}\) branch and links it with biomarker sensitivity through \(X_{34}\).

\subsection{Pathway modelling in biomedical data}
Causal graphical models and mediation analysis provide tools for decomposing exposure--mediator--outcome relationships \cite{pearl2009causality,imai2010mediation}. Generalised additive models provide a flexible way to represent nonlinear biomedical relationships while retaining interpretable smooth terms \cite{hastie1990gam,wood2017gam}. In this paper, these ideas are used in a conservative way: $X1234$ supports pathway ranking, whereas NHANES $R^*$ supports participant-level pathway summaries. The goal is pathway prioritisation and biomedical hypothesis generation rather than proof of biological causality. Unlike standard mediation models that require all variables in one cohort, Causal-RetiGraph separates
cross-cohort pathway prioritisation through \(X1234\) from same-subject pathway summaries through \(R^*\).

\section{Methodology}

\subsection{Overview of the Causal-RetiGraph workflow}

Fig.~\ref{fig:workflow} summarises the complete Causal-RetiGraph workflow. The framework is organised as a sequential retinal--systemic analysis pipeline instead of a standalone DR classifier. The workflow contains nine connected steps. Steps 1--4 form the retinal-image support fold, where an external DR-retinal Cohort is used to construct, train, validate and test the image-derived retinal graph phenotype denoted \(X1234\). Steps 5--8 form the systemic pathway fold, where NHANES 2005--2008 participant-level data are used to anchor systemic exposures, estimate binary DR associations, prioritise candidate pathways using \(X1234\)-based retinal support, and perform participant-level pathway analysis using the same-subject NHANES retinal mediator family \(R^*\). Step 9 summarises the main outputs of the framework.
\begin{table*}[t]
\centering
\caption{Data provenance and analytical roles in Causal-RetiGraph.}
\label{tab:roles}
\small
\renewcommand{\arraystretch}{1.15}

\begin{tabular*}{\textwidth}{@{\extracolsep{\fill}}
p{0.12\textwidth}
p{0.34\textwidth}
p{0.48\textwidth}}
\hline\hline
\textbf{Construct} & \textbf{Source and role} & \textbf{Interpretation} \\
\hline
\(X_1\) &
Retinal cohort; vessel stream &
Artery, vein and vessel segmentation \\
\hline
\(X_2\) &
Retinal cohort; lesion-evidence stream &
Grad-CAM heatmap and lesion-summary evidence \\
\hline
\(X_3\) &
Retinal cohort; embedding stream &
Lesion-sensitive representation vector used in the \(X_{34}\) Jacobian branch \\
\hline
\(X_4\) &
Retinal cohort; biomarker stream &
AutoMorph biomarkers or tabular oculomics \\
\hline
\(X_{1234}\) &
Retinal cohort; fused retinal graph phenotype &
Held-out binary and graded DR evaluation \\
\hline
\(E\) &
NHANES; exposure block &
Glycaemic, renal, haemodynamic, lipid and inflammatory variables \\
\hline
\(Z,G\) &
NHANES; adjustment blocks &
Baseline covariates and ancestry-proxy/sensitivity variables \\
\hline
\(R^*\) &
NHANES; same-subject retinal mediator family &
Participant-level retinal/context family summarised by \(\mathrm{Rstar\_family\_pca1}\) \\
\hline
\(Y\) &
NHANES; outcome block &
Binary DR and renal, haemodynamic or inflammatory outcome \\
\hline\hline
\end{tabular*}
\end{table*}
The workflow separates retinal-image evaluation from participant-level systemic pathway modelling. The fused retinal phenotype \(X1234\) is learned and evaluated within the external DR-XAI image cohort using train--validation--test splits. This establishes \(X1234\) as a validated image-derived retinal support phenotype for DR discrimination and severity-sensitive representation. NHANES then provides an independent systemic anchoring source. In NHANES, systemic exposures, covariates, retinal/context variables and outcomes are measured at the participant level through SEQN-linked records. Therefore, \(X1234\) is used as retinal support for cross-cohort pathway prioritisation, whereas \(R^*\) is used for same-subject pathway summaries within NHANES. This distinction avoids overclaiming that the externally derived \(X1234\) phenotype has been directly measured on the same NHANES participants. Table \ref{tab:roles} provides an overview of all constructs used in the framework, their data provenance, analytical roles, and interpretations.

\subsection{Steps 1--2: Retinal image cohort and dual-edge evidence streams}

The retinal-image support fold begins with a curated DR image cohort derived from the APTOS 2019 Blindness Detection dataset \cite{aptos2019-blindness-detection}. This cohort is internal to the retinal-image fold but external to the NHANES participant-level pathway fold. We use a carefully curated, non-augmented subset of 2,910 fundus images with image-level DR grades from 0 to 4. The data are split using a stratified 60/20/20 protocol into 1,746 training images, 582 validation images and 582 test images. This split is used to train, validate and test the image-derived retinal graph phenotype \(X1234\) before it is used as retinal support evidence in the cross-cohort pathway analysis.

Each retained image is represented by four aligned evidence streams. The first stream, \(X_1\), captures segmentation-derived retinal vascular anatomy, comprising artery, vein, whole-vessel, and macular-region maps that preserve vessel topology and spatial geometry. The second stream, \(X_2\), comprises lesion evidence maps and lesion-summary statistics derived from weak DR-XAI localisation. In this study, \(X_2\) is interpreted as class-discriminative lesion evidence in lieu of expert-annotated lesion segmentation. Grad-CAM-style localisation is used to obtain image regions contributing to DR prediction \cite{selvaraju2017}. The third stream, \(X_3\), is a lesion-sensitive image embedding. A Huang-style lesion-based contrastive ResNet50 checkpoint is used to obtain high-dimensional image features, which are then projected into the canonical 128-dimensional embedding schema \cite{huang2021}. The fourth stream, \(X_4\), contains AutoMorph-derived retinal morphometric biomarkers, including vessel density, tortuosity, average vessel width, fractal dimension, CRAE, CRVE and AVR \cite{Zhou2022AutoMorph}.

The stream order is fixed to avoid ambiguity between lesion evidence and image embeddings. In particular, \(X_2\) denotes the spatial lesion-evidence stream used with vessel maps in the \(X_{12}\) spatial branch, whereas \(X_3\) denotes the vector embedding stream used in the \(X_{34}\) Jacobian branch. This distinction is important because the two streams serve different roles: \(X_2\) preserves local image-space evidence, while \(X_3\) provides a compact representation for embedding--biomarker sensitivity analysis.

\subsection{Step 3: Dual-branch representation and fusion}

The four streams are organised into two complementary branches. The first branch is the spatial branch, denoted \(X_{12}\). It combines vessel structure \((X_1)\) and lesion evidence \((X_2)\) in image space to model lesion--vessel geometry. Let \(A_i\), \(R_i\) and \(V_i\) denote artery, vein and whole-vessel maps for image \(i\), let \(L_i\) denote lesion evidence, and let \(Z_i\) denote macular-zone information. The spatial interaction channels are defined as
\begin{equation}
M_i^{12}=\{A_i,R_i,V_i,L_i,L_i\odot A_i,L_i\odot R_i,L_i\odot V_i,Z_i\},
\end{equation}
where \(\odot\) denotes pixel-wise interaction. A spatial representation \(z_i^{12}\) is extracted from these channels, and image similarity in the spatial branch is computed as
\begin{equation}
E^{12}_{ij}=\cos(z_i^{12},z_j^{12}).
\end{equation}
This edge family measures whether two images share similar lesion--vessel geometry rather than only similar DR labels.

The second branch is the Jacobian branch, denoted \(X_{34}\). It links the lesion-sensitive embedding stream \((X_3)\) with the biomarker stream \((X_4)\). A differentiable mapper is learned from the embedding to a reduced biomarker representation:
\begin{equation}
\hat{x}_{4,i}=f_{\theta}(x_{3,i}),\qquad
J_i=\frac{\partial f_{\theta}(x_{3,i})}{\partial x_{3,i}}.
\end{equation}
The Jacobian \(J_i\) summarises representation-space sensitivity between image embeddings and retinal biomarkers. The \(X_{34}\) descriptor includes reduced \(X_3\), reduced \(X_4\), the Frobenius norm of \(J_i\), and input/output sensitivity summaries. The corresponding branch similarity is
\begin{equation}
E^{34}_{ij}=\cos(z_i^{34},z_j^{34}).
\end{equation}
The Jacobian descriptor is used as an interpretability signal and is not interpreted as biological causality.

The two branches are fused using two-token attention:
\begin{equation}
h_i=\alpha_i^{12}\phi_{12}(z_i^{12})+\alpha_i^{34}\phi_{34}(z_i^{34}),
\end{equation}
where \(\phi_{12}\) and \(\phi_{34}\) are branch-specific projection functions, and \(\alpha_i^{12}\) and \(\alpha_i^{34}\) are attention weights. The resulting fused retinal graph phenotype is denoted \(X1234\).
\begin{table}[t]
\centering
\caption{APTOS-DR5 class test performance (\(n=582\)).}
\label{tab:aptos_dr5}
\scriptsize
\renewcommand{\arraystretch}{1.12}

\resizebox{\columnwidth}{!}{%
\begin{tabular}{lccccc}
\hline\hline
\textbf{Stream/model} & \textbf{Acc.} & \textbf{QWK} & \textbf{Macro-F1} &
\textbf{MAE} & \textbf{Adj. Acc.} \\
\hline
\(X_1\) vessel & .6409 & .5038 & .2832 & .6186 & .7887 \\
\(X_2\) lesion evidence & .7388 & .6720 & .4256 & .4244 & .8729 \\
\(X_3\) Huang LCL & .8110 & .8265 & .5934 & .2749 & .9278 \\
\(X_4\) biomarkers & .6770 & .5809 & .3553 & .5498 & .8076 \\
\(X_{12}\) spatial & .7354 & .6890 & .4258 & .4124 & .8832 \\
\(X_{34}\) Jacobian & .8007 & .8280 & .5826 & .2835 & .9296 \\
Full graph \(X_{1234}\) & .8076 & .8312 & .5915 & .2749 & .9330 \\
\hline\hline
\vspace{-0.55cm}
\end{tabular}}
\end{table}

\subsection{Step 4: External evaluation of \texorpdfstring{$X1234$}{X1234}}

Step 4 evaluates the fused retinal graph phenotype \(X1234\) within the external DR-XAI retinal cohort. In this step, the retinal model is trained, validated and tested on the DR image dataset using held-out evaluation. Binary evaluation measures the ability of \(X1234\) to separate no-DR from DR grades 1--4, while graded evaluation measures severity-sensitive performance across DR grades 0--4. The reported metrics include AUROC, AUPRC, quadratic weighted kappa and adjacent-grade accuracy.

In the workflow, this step is labelled as external evaluation because the DR-XAI retinal cohort is external to the NHANES systemic pathway dataset. It does not mean that \(X1234\) is directly measured or externally validated on the same NHANES participants. Instead, Step 4 establishes that \(X1234\) captures meaningful lesion--vascular and biomarker-sensitive retinal structure within an independent retinal-image cohort. The arrow from Step 4 to Step 5 therefore represents analytical anchoring: the evaluated \(X1234\) phenotype provides retinal support evidence that is carried forward into cross-cohort pathway prioritisation, while NHANES provides the independent systemic exposure and outcome anchor.
\subsection{Step 5: NHANES 2005--2008 anchoring}

The systemic fold uses NHANES 2005--2008 diabetic participants with retinal photographs linked to systemic health and laboratory variables. Participants are linked through SEQN identifiers. This fold provides the same-subject epidemiological anchor for systemic exposure--DR analysis. The NHANES subset includes binary DR status, systemic exposures, covariates, retinal/context variables and outcome-family variables. 

Let \(E\) denote systemic exposures, \(Z\) baseline covariates, \(G\) an ancestry-proxy or sensitivity block, \(R^*\) the NHANES retinal mediator family, and \(Y\) binary DR or outcome-family variables. Table~\ref{tab:roles} summarises the provenance, definitions, and analytical roles of these constructs. In contrast to \(X1234\), which is obtained from the external retinal-image cohort, \(R^*\) is constructed within NHANES and is therefore available at the same participant level as \(E\), \(Z\), \(G\), and \(Y\). This makes \(R^*\) the appropriate retinal construct for same-subject pathway summaries.

\subsection{Step 6: NHANES binary DR association analysis}
\begin{table}[t]
\centering
\caption{Binary DR test performance for \(X_{1234}\) phenotype.}
\label{tab:binary_dr}
\scriptsize
\renewcommand{\arraystretch}{1.12}

\resizebox{\columnwidth}{!}{%
\begin{tabular}{lcccccc}
\hline\hline
\textbf{Model} & \textbf{Acc.} & \textbf{Bal. Acc.} & \textbf{Sens.} &
\textbf{Spec.} & \textbf{AUROC} & \textbf{AUPRC} \\
\hline
\(X_{1234}\) & .9055 & .9038 & .8964 & .9111 & .9711 & .9423 \\
\hline\hline
\vspace{-1.0cm}
\end{tabular}}
\end{table}

For each systemic exposure \(E_j\), the binary DR anchor estimates the association between systemic variation and DR status:
\begin{equation}
\logit P(Y_{DR}=1)=\alpha+\beta_jE_j+\gamma^\top Z+\delta^\top G.
\end{equation}
Here, \(\beta_j\) represents the adjusted exposure--DR association, while \(Z\) and \(G\) account for baseline covariates and sensitivity variables. This step identifies systemic factors most strongly associated with DR in the NHANES diabetic subset.

In the workflow, example exposures include HbA1c, urine albumin and pulse pressure. These variables represent glycaemic burden, renal microvascular injury and haemodynamic stress, respectively. This stage defines the systemic association side of the framework. It estimates exposure--DR association strength only and does not by itself estimate mediation.

\subsection{Step 7: Cross-cohort pathway prioritisation with X1234 support}

The cross-cohort prioritisation layer combines NHANES exposure--DR association strength with retinal support from \(X1234\). For each exposure family \(j\), the indirect-priority index is defined as
\begin{equation}
P_j=\mathrm{norm}\{|\hat\beta_j|\times S_j^{X1234}\times[-\log_{10}(q_j)]\},
\end{equation}
where \(\hat\beta_j\) is the NHANES exposure--DR coefficient, \(q_j\) is the FDR-adjusted association value, and \(S_j^{X1234}\) is the retinal support score derived from the externally evaluated image-based phenotype. The support score reflects whether the retinal graph phenotype contains lesion--vascular or biomarker-sensitive evidence consistent with the pathway family.

This score ranks candidate exposure--retina--DR pathways. For example, HbA1c-linked pathways are interpreted as glycaemic lesion--vascular pathways, pulse-pressure-linked pathways as haemodynamic stress pathways, and albuminuria-linked pathways as renal microvascular injury pathways. The priority score is not a natural indirect effect and is not interpreted as participant-level mediation. It is a cross-source ranking index that combines systemic evidence from NHANES with retinal support from \(X1234\). Its purpose is to identify biologically plausible pathway families for future same-subject testing.

\subsection{Participant-level R* pathway summaries}

Participant-level pathway analysis is performed within NHANES using the same-subject retinal mediator family:
\begin{equation}
R^*=\mathrm{Rstar\_family\_pca1}.
\end{equation}
The pathway structure is defined as
\begin{equation}
E\rightarrow R^*\rightarrow Y\mid Z,G.
\end{equation}
Because \(R^*\), \(E\), \(Z\), \(G\) and \(Y\) are all defined within the NHANES participant-level dataset, this step is the appropriate part of the framework for TE-, NDE- and NIE-style summaries. These summaries are estimated using bootstrap confidence intervals and FDR correction.

The pathway quantities are described as mediation-style summaries rather than definitive causal effects because they depend on model specification, cross-sectional assumptions and the available NHANES variables. This step is therefore used for hypothesis generation about whether retinal variables behave as downstream correlates, mediator-like signals or structured intermediate markers in diabetic microvascular pathways.

\begin{figure*}[t]
\centering
\begin{subfigure}[t]{0.48\textwidth}
\centering
\includegraphics[width=\linewidth]{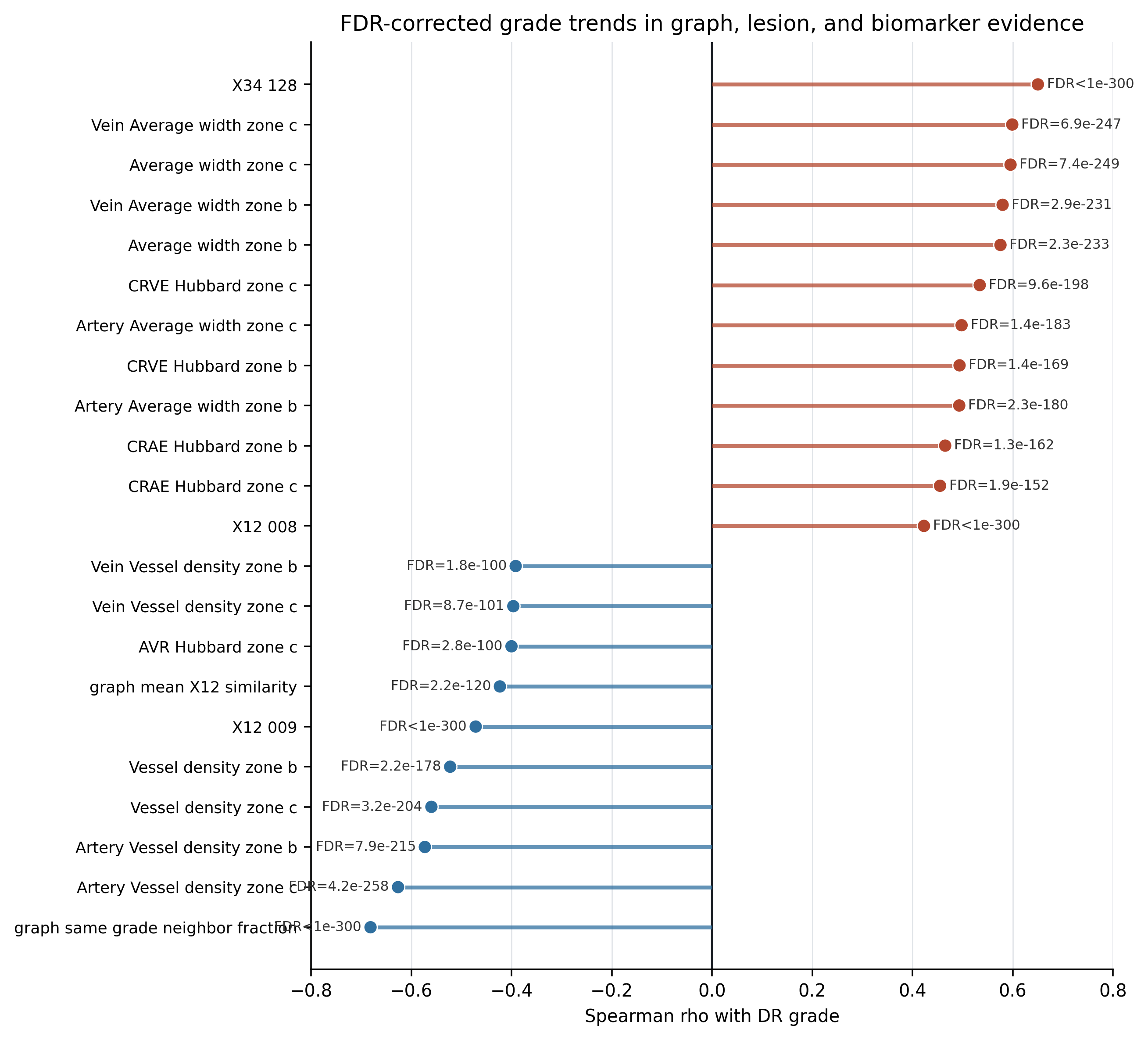}
\caption{FDR-corrected grade trends.}
\end{subfigure}
\hfill
\begin{subfigure}[t]{0.48\textwidth}
\centering
\includegraphics[width=\linewidth]{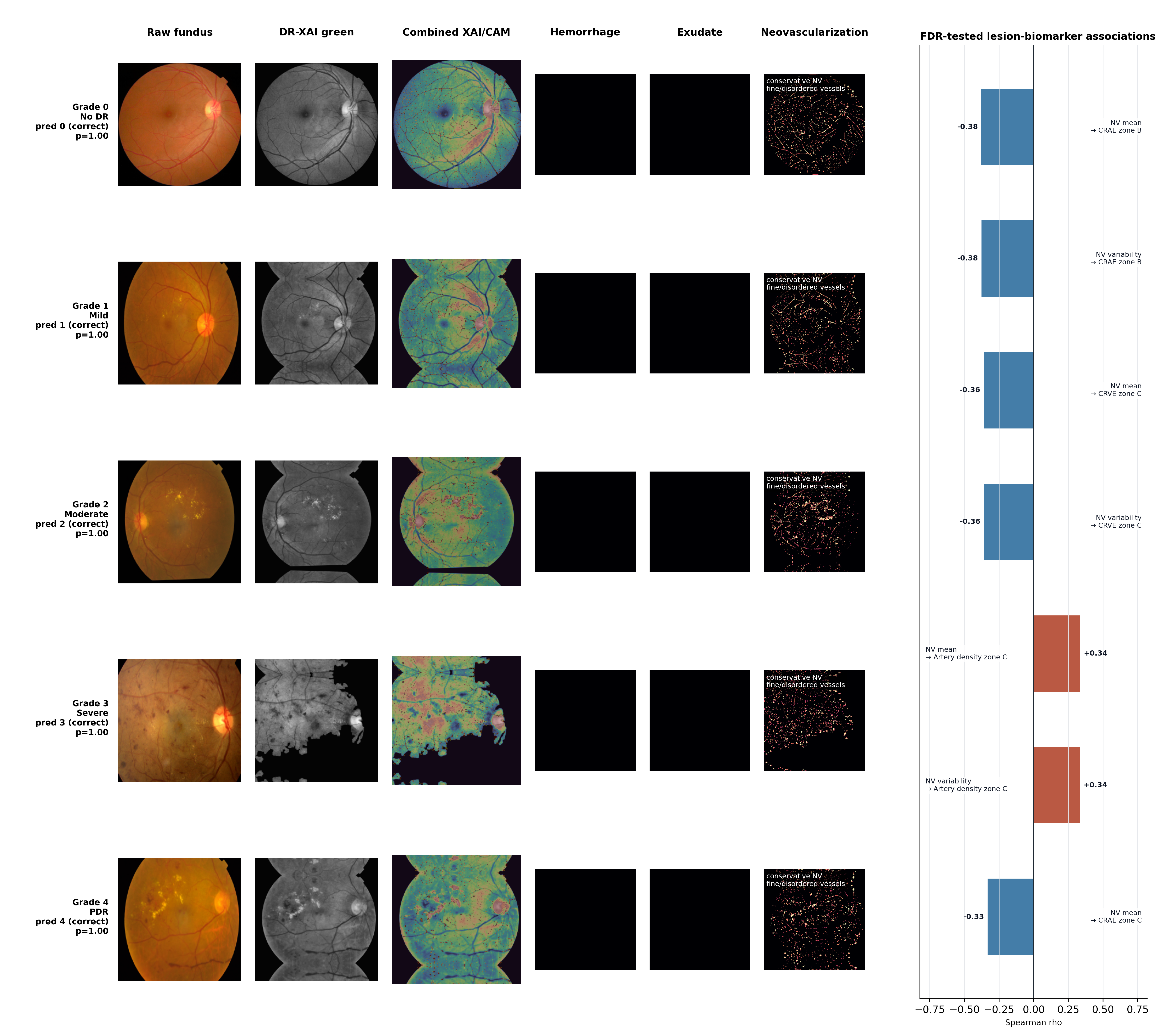}
\caption{FDR-corrected lesion--biomarker associations.}
\end{subfigure}

\caption{Retinal support evidence for \(X1234\).
(a) FDR-corrected grade-trend analysis showing severity-sensitive graph, lesion and biomarker components
across DR grades. Positive associations indicate features increasing with DR severity, while negative
associations indicate features decreasing with severity.
(b) FDR-corrected lesion--biomarker associations showing that weakly supervised lesion evidence is coupled
with AutoMorph vascular measurements. Together, the panels support the interpretation that \(X1234\)
captures disease-severity structure and lesion--vascular coupling, rather than only image-level
classification performance.}
\label{fig:retinal_support}
\end{figure*}


\section{Results}
\subsection{Retinal support evidence}
The retinal-image validation results indicate that \(X_{1234}\) captures clinically relevant DR-related information across both disease detection and severity assessment. For binary DR classification, \(X_{1234}\) achieved an accuracy of 0.9055, balanced accuracy of 0.9038, sensitivity of 0.8964, specificity of 0.9111, AUROC of 0.9711, and AUPRC of 0.9423 (Table \ref{tab:binary_dr}). For five-class DR grading, performance reached an accuracy of 0.8076, QWK of 0.8312, macro-F1 score of 0.5915, and adjacent-grade accuracy of 0.9330 (Table \ref{tab:aptos_dr5}).

Fig \ref{fig:retinal_support} further supports the biological relevance of the learned phenotype. As DR severity increases, calibre- and Jacobian-related features show increasing trends, whereas vessel-density and graph-neighbourhood features generally decline (Fig.~\ref{fig:retinal_support}a). These patterns suggest that \(X_{1234}\) reflects progressive microvascular alterations associated with disease advancement. In addition, lesion-evidence features demonstrate consistent associations with AutoMorph-derived vascular biomarkers (Fig.~\ref{fig:retinal_support}b), supporting the integration of lesion, vascular, and biomarker information through the \(X_{12}\) and \(X_{34}\) graph branches. Collectively, these findings support the use of \(X_{1234}\) as a severity-sensitive retinal phenotype for cross-cohort pathway prioritization.

\subsection{NHANES association layer and cross-cohort prioritisation}
Fig.~\ref{fig:association} shows the NHANES binary DR association layer. The strongest adjusted systemic
factors are HbA1c, urine albumin, pulse pressure, fasting glucose and systolic blood pressure, placing
glycaemic burden first, followed by renal microvascular stress and haemodynamic load. These results define
the systemic exposure--DR side of the framework before retinal support is introduced.
Table~\ref{tab:x1234_priority} reports the top five cross-cohort \(X1234\)-centred pathway priorities. The
association terms \(\beta\), OR and FDR are estimated from the NHANES binary DR anchor, whereas the
priority score combines this systemic association strength with \(X1234\)-based retinal support. HbA1c
produces the highest-ranked pathway, followed by pulse pressure, fasting glucose, urine albumin and
systolic blood pressure. These pathways should be interpreted as prioritisation hypotheses, not
participant-level mediation effects.

\subsection{NHANES anchor and prioritisation}
\begin{figure*}[t]
\centering
\includegraphics[width=\textwidth]{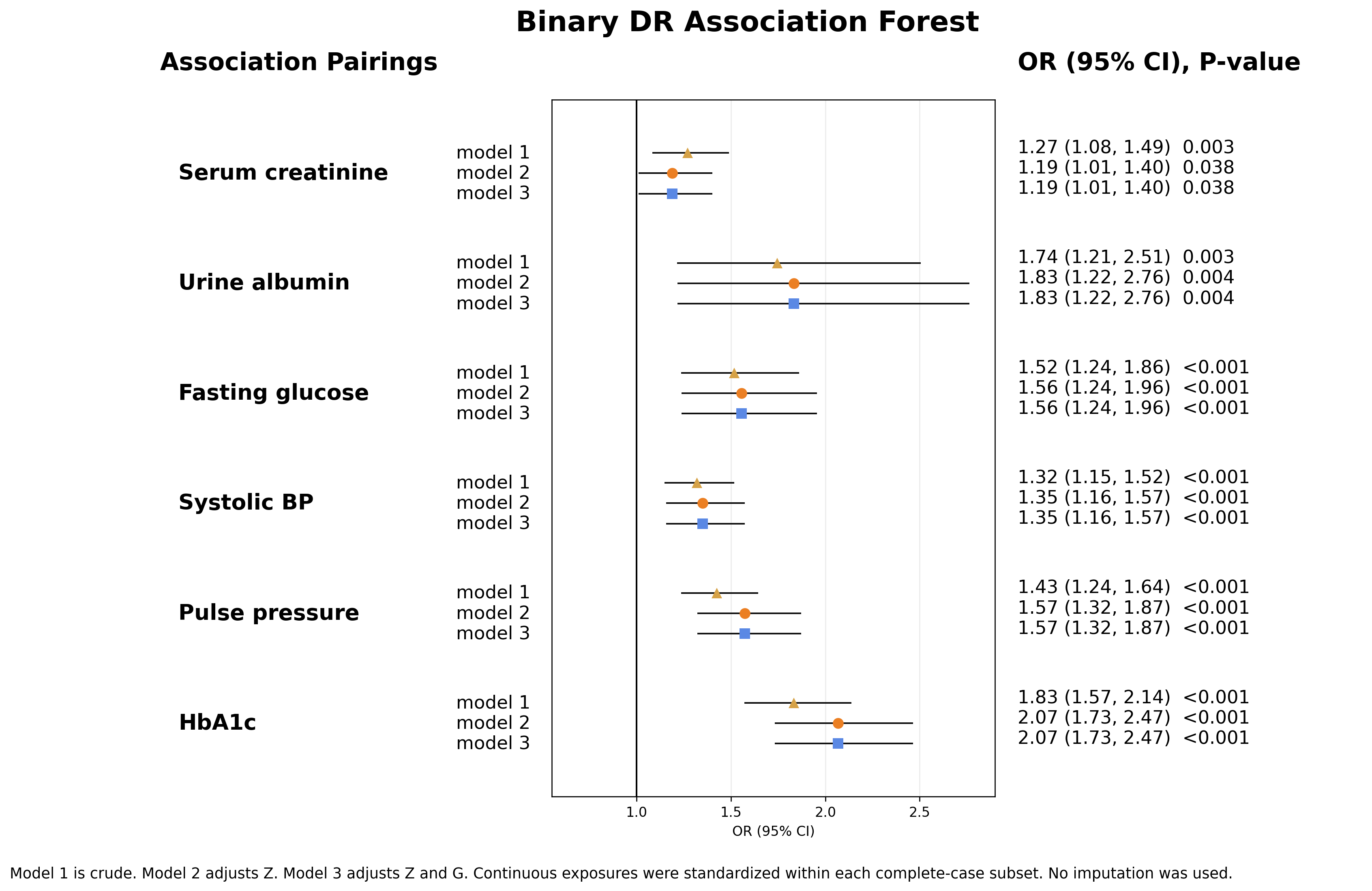}
\caption{NHANES binary DR association anchor in the 884-row diabetes subset. Model 1 is crude, Model 2 adjusts for $Z$, and Model 3 adjusts for $Z+G$. The strongest adjusted factors are HbA1c, urine albumin, pulse pressure, fasting glucose and systolic blood pressure.}
\label{fig:association}
\end{figure*}

The strongest adjusted NHANES binary DR anchors are HbA1c (OR 2.0669), urine albumin (OR 1.8335), pulse pressure (OR 1.5734), fasting glucose (OR 1.5566) and systolic blood pressure (OR 1.3489). These results place glycaemic burden first, followed by renal microvascular stress and haemodynamic load. Figure~\ref{fig:association} shows the observed systemic side of the framework before any retinal prioritisation is applied.
Cross-source prioritisation ranks HbA1c $\rightarrow X1234 \rightarrow$ glycaemic lesion--vascular burden as the leading candidate pathway, followed by fasting glucose, pulse pressure, urine albumin and systolic blood pressure. The prioritisation figure should be read as a decision-support layer for model development: it combines a systemic association signal with an image-derived retinal support score, but it does not estimate a participant-level indirect effect.
\begin{table*}[t]
\centering
\caption{Top five cross-cohort \(X_{1234}\)-centred pathway priorities.}
\label{tab:x1234_priority}
\small
\renewcommand{\arraystretch}{1.15}

\begin{tabular*}{\textwidth}{@{\extracolsep{\fill}}
p{0.60\textwidth}cccc}
\hline\hline
\textbf{Pathway} & \(\boldsymbol{\beta}\) & \textbf{OR} & \textbf{FDR} & \textbf{Priority} \\
\hline
LBXGH \(\rightarrow X_{1234} \rightarrow\) glycaemic lesion--vascular pathway
& .7260 & 2.0669 & \(1.16\times10^{-14}\) & 10.1186 \\

PulsePressure\_mean \(\rightarrow X_{1234} \rightarrow\) haemodynamic vascular stress
& .4533 & 1.5734 & \(2.88\times10^{-6}\) & 2.5114 \\

LBXGLU \(\rightarrow X_{1234} \rightarrow\) glycaemic lesion--vascular pathway
& .4425 & 1.5566 & \(6.25\times10^{-4}\) & 1.4179 \\

URXUMA \(\rightarrow X_{1234} \rightarrow\) albuminuria microvascular injury
& .6062 & 1.8335 & \(8.03\times10^{-3}\) & 1.2701 \\

SBP\_mean \(\rightarrow X_{1234} \rightarrow\) blood-pressure vascular stress
& .2993 & 1.3489 & \(6.25\times10^{-4}\) & .9588 \\
\hline\hline
\end{tabular*}

\vspace{1mm}

\begin{minipage}{\textwidth}
\footnotesize
\emph{Note:} \(\beta\), OR and FDR are estimated from the NHANES binary DR anchor. Priority is the cross-cohort ranking score combining NHANES exposure--DR association strength with \(X_{1234}\)-based retinal support. These pathways are prioritisation hypotheses, not participant-level mediation effects. LBXGH = HbA1c; LBXGLU = fasting glucose; URXUMA = urine albumin; SBP = systolic blood pressure.
\end{minipage}
\end{table*}
\subsection{Participant-level R* pathway summaries}
\begin{figure*}[t]
\centering
\includegraphics[width=0.95\textwidth]{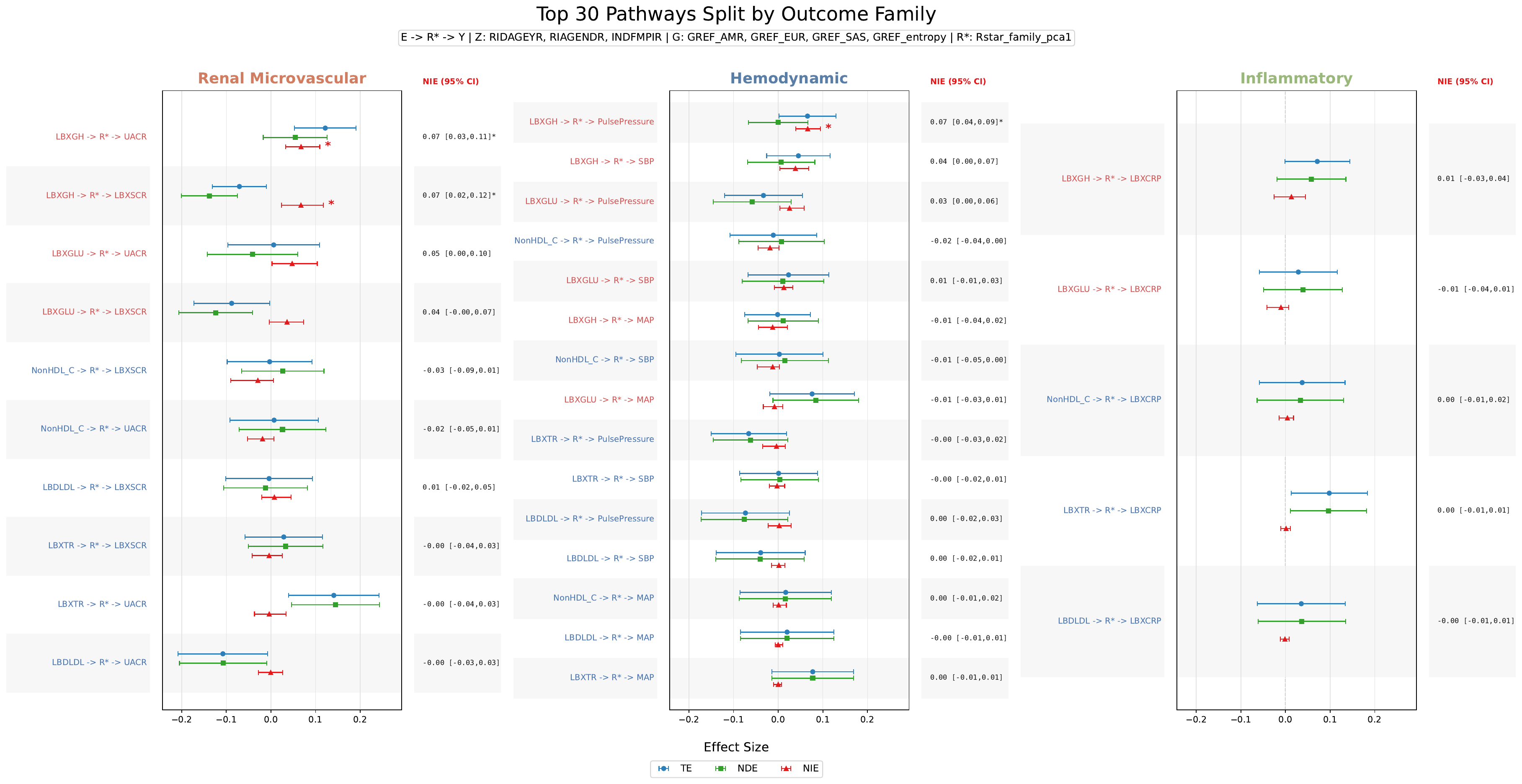}
\caption{Participant-level NHANES pathway summaries split by outcome family. The mediator is the
same-subject NHANES retinal mediator family \(R^*=\mathrm{Rstar\_family\_pca1}\), not the image-derived
\(X1234\) phenotype. Renal microvascular, haemodynamic and inflammatory outcome families are shown
separately. Points and intervals indicate TE-, NDE- and NIE-style summaries with bootstrap confidence
intervals. The strongest mediator-style signals occur in glycaemic--renal and glycaemic--haemodynamic
pathways, while inflammatory pathways show weaker and less consistent patterns.}
\label{fig:pathways}
\end{figure*}

Fig.~\ref{fig:pathways} reports the same-subject NHANES pathway analysis using \(R^*\). The renal panel shows the clearest mediator-style signals, especially for HbA1c-linked pathways to UACR and serum creatinine. The haemodynamic panel also shows a visible HbA1c-linked pathway to pulse pressure. In contrast, pathways involving C-reactive protein (CRP), a common biomarker of systemic inflammation, exhibit weaker and less consistent effects. These results
suggest that the NHANES \(R^*\) family aligns more strongly with glycaemic--renal and
glycaemic--haemodynamic axes than with broad inflammation. Importantly, these are \(R^*\)-based
same-subject pathway summaries and should not be interpreted as \(E\rightarrow X1234\rightarrow Y\)
mediation.

\section{Discussion}
Causal-RetiGraph provides a compact biomedical informatics framework for connecting retinal image evidence
with systemic pathway analysis in DR. The main finding is the recurrence of glycaemic, renal and
haemodynamic axes across the retinal-image and NHANES folds. In the retinal fold, \(X1234\) shows strong
binary and graded DR performance and captures severity-sensitive lesion--vascular structure. In the NHANES
fold, HbA1c, urine albumin, pulse pressure, fasting glucose and systolic blood pressure dominate the
systemic DR association landscape. In the participant-level pathway fold, \(R^*\)-based summaries
highlight glycaemic--renal and glycaemic--haemodynamic pathways more clearly than inflammatory pathways.
This pattern supports the interpretation that retinal graph phenotypes can provide a structured
intermediate layer between local retinal injury and systemic diabetic microvascular stress. The framework
does not claim that \(X1234\) is a measured NHANES mediator. Instead, \(X1234\) provides image-derived
retinal support for pathway prioritisation, while \(R^*\) provides the same-subject retinal construct for
NHANES pathway summaries. This distinction is central to the validity of the study.

\subsection{Glycaemic, renal and haemodynamic axes}
The recurrence of these pathway families is the main result. HbA1c and fasting glucose define the glycaemic axis. Urine albumin and serum creatinine define the renal microvascular axis. Pulse pressure and systolic blood pressure define the haemodynamic axis. These axes are biologically connected: hyperglycaemia damages microvascular structure, renal dysfunction reflects systemic microvascular injury and haemodynamic stress can influence vessel calibre and vascular remodelling. Retinal imaging provides a measurable location where these processes may become visible as lesion burden, calibre change, density change and graph-neighbourhood disruption. The ranking in Table~\ref{tab:x1234_priority} is consistent with Fig.~\ref{fig:association}, where glycaemic, renal and haemodynamic variables dominate the adjusted systemic association landscape.

\subsection{Interpretability beyond classification}
The retinal branch does not only improve prediction. It decomposes retinal evidence into spatial and vector-space relations. The spatial $X_{12}$ branch asks where lesion-like evidence occurs relative to vessels and macular zones. The Jacobian $X_{34}$ branch asks which retinal biomarker directions are sensitive to lesion/image embedding changes. This separation makes the representation useful for biomedical interpretation: a model output can be inspected through severity trends, lesion--biomarker links, branch-level sensitivity and graph topology, rather than through a single probability score.The patterns in Fig.~\ref{fig:retinal_support} show why the retinal fold is interpretable beyond
classification: \(X_{12}\) preserves lesion--vessel geometry, while \(X_{34}\) links embedding changes to
biomarker sensitivity.

\subsection{Why X1234 and R* must remain distinct}
The distinction between $X1234$ and $R^*$ is essential. $X1234$ is an image-derived support phenotype and is used for cross-source prioritisation. $R^*$ is an NHANES same-subject mediator family and is used for participant-level pathway summaries. The study therefore does not claim $E\rightarrow X1234\rightarrow Y$ mediation in NHANES. Instead, it provides a sequence: define interpretable retinal phenotypes, use them to prioritise systemic pathways and then test the corresponding mediator structure in same-subject data. This is the safest way to combine image-derived retinal AI with population-scale biomedical data without overstating causal claims. Fig.~\ref{fig:pathways} further supports this separation by showing that participant-level pathway summaries are estimated through \(R^*\), not through \(X1234\).

\subsection{Biomedical informatics contribution}
The main contribution of Causal-RetiGraph is to connect retinal image evidence with participant-level systemic pathway analysis while preserving data provenance. The retinal fold converts lesion, vessel, embedding and biomarker information into the interpretable \(X1234\) graph phenotype. The NHANES fold anchors systemic exposure--DR associations and uses \(R^*\) for same-subject pathway summaries. This structure allows retinal AI outputs to be used as biomedical evidence rather than only as classification scores.
\section{Limitations}
Several limitations define the current scope. First, $X1234$ is not observed on the same NHANES participants, so the prioritisation analysis is not a participant-level mediation estimate. Second, $R^*$ is an NHANES retinal grading/context mediator family and may not contain all information in image-derived $X1234$. Third, lesion evidence is weak DR-XAI-style evidence rather than expert lesion segmentation. Fourth, the nonlinear smooth branch is exploratory because the linear branch currently gives the more stable pattern. 

\section{Conclusion}
Causal-RetiGraph links retinal graph phenotypes, NHANES systemic anchoring and same-subject pathway modelling for diabetic retinopathy. The framework shows that \(X1234\) provides interpretable retinal support, NHANES anchors systemic exposure--DR associations, and \(R^*\) supports participant-level pathway summaries. Across these folds, glycaemic, renal and haemodynamic axes consistently dominate. The next step is same-subject retinal-systemic modelling in which $X1234$ is computed directly on SEQN-linked retinal images and $E\rightarrow X1234\rightarrow Y$ can be estimated without cross-source bridging.









\bibliographystyle{IEEEtran}
\bibliography{BIBM_2026_Causal_RetiGraph_References}

\begin{thebibliography}{10}
\providecommand{\url}[1]{#1}
\csname url@samestyle\endcsname
\providecommand{\newblock}{\relax}
\providecommand{\bibinfo}[2]{#2}
\providecommand{\BIBentrySTDinterwordspacing}{\spaceskip=0pt\relax}
\providecommand{\BIBentryALTinterwordstretchfactor}{4}
\providecommand{\BIBentryALTinterwordspacing}{\spaceskip=\fontdimen2\font plus
\BIBentryALTinterwordstretchfactor\fontdimen3\font minus \fontdimen4\font\relax}
\providecommand{\BIBforeignlanguage}[2]{{%
\expandafter\ifx\csname l@#1\endcsname\relax
\typeout{** WARNING: IEEEtran.bst: No hyphenation pattern has been}%
\typeout{** loaded for the language `#1'. Using the pattern for}%
\typeout{** the default language instead.}%
\else
\language=\csname l@#1\endcsname
\fi
#2}}
\providecommand{\BIBdecl}{\relax}
\BIBdecl

\bibitem{antar2023diabetes}
S.~A. Antar, N.~A. Ashour, M.~Sharaky, M.~Khattab, N.~A. Ashour, R.~T. Zaid, E.~J. Roh, A.~Elkamhawy, and A.~A. Al-Karmalawy, ``Diabetes mellitus: Classification, mediators, and complications; a gate to identify potential targets for the development of new effective treatments,'' \emph{Biomedicine \& Pharmacotherapy}, vol. 168, p. 115734, 2023.

\bibitem{giannakogeorgou2026diabetes}
A.~Giannakogeorgou, M.~Roden, and K.~Pafili, ``Diabetes mellitus as a multisystem disease: understanding subtypes, complications, and the link with steatotic liver diseases in humans,'' \emph{Hormones}, vol.~25, no.~1, pp. 61--80, 2026.

\bibitem{WHO2024Diabetes}
{World Health Organization}, ``Diabetes,'' \url{https://www.who.int/news-room/fact-sheets/detail/diabetes}, Nov. 2024, fact sheet. Accessed 2026-07-03.

\bibitem{Lu2023Vascular}
Y.~Lu, W.~Wang, J.~Liu, M.~Xie, Q.~Liu, and S.~Li, ``Vascular complications of diabetes: A narrative review,'' \emph{Medicine}, vol. 102, no.~40, p. e35285, 2023.

\bibitem{Zakir2023CardiovascularDiabetes}
M.~Zakir, N.~Ahuja, M.~A. Surksha, R.~Sachdev, Y.~Kalariya, M.~Nasir, M.~Kashif, F.~Shahzeen, A.~Tayyab, M.~S.~M. Khan, M.~Junejo, F.~M. Kumar, G.~Varrassi, S.~Kumar, M.~Khatri, and T.~Mohamad, ``Cardiovascular complications of diabetes: From microvascular to macrovascular pathways,'' \emph{Cureus}, vol.~15, no.~9, p. e45835, 2023.

\bibitem{Kulkarni2024ClinicalRelationship}
A.~Kulkarni, A.~R. Thool, and S.~Daigavane, ``Understanding the clinical relationship between diabetic retinopathy, nephropathy, and neuropathy: A comprehensive review,'' \emph{Cureus}, vol.~16, no.~3, p. e56674, 2024.

\bibitem{wei2022pathophysiological}
L.~Wei, X.~Sun, C.~Fan, R.~Li, S.~Zhou, and H.~Yu, ``The pathophysiological mechanisms underlying diabetic retinopathy,'' \emph{Frontiers in Cell and Developmental Biology}, vol.~10, p. 963615, 2022.

\bibitem{zhu2025oculomics}
Z.~Zhu, Y.~Wang, Z.~Qi, W.~Hu, X.~Zhang, S.~K. Wagner, Y.~Wang, A.~R. Ran, J.~Ong, E.~Waisberg \emph{et~al.}, ``Oculomics: Current concepts and evidence,'' \emph{Progress in Retinal and Eye Research}, vol. 106, p. 101350, 2025.

\bibitem{Chew2025}
E.~Y. Chew, S.~A. Burns, A.~G. Abraham \emph{et~al.}, ``Standardization and clinical applications of retinal imaging biomarkers for cardiovascular disease: A roadmap from an {NHLBI} workshop,'' \emph{Nature Reviews Cardiology}, vol.~22, no.~1, pp. 47--63, 2025.

\bibitem{Zhang2024}
Z.~Zhang, C.~Deng, and Y.~M. Paulus, ``Advances in structural and functional retinal imaging and biomarkers for early detection of diabetic retinopathy,'' \emph{Biomedicines}, vol.~12, no.~7, p. 1405, 2024.

\bibitem{Gulshan2016}
V.~Gulshan, L.~Peng, M.~Coram, M.~C. Stumpe, D.~Wu, A.~Narayanaswamy, S.~Venugopalan, K.~Widner, T.~Madams, J.~Cuadros \emph{et~al.}, ``Development and validation of a deep learning algorithm for detection of diabetic retinopathy in retinal fundus photographs,'' \emph{JAMA}, vol. 316, no.~22, pp. 2402--2410, 2016.

\bibitem{Nadeem2022DRReview}
M.~W. Nadeem, H.~G. Goh, V.~Ponnusamy, I.~Andonovic, M.~A. Khan, and M.~Hussain, ``Deep learning for diabetic retinopathy analysis: A review, research challenges, and future directions,'' \emph{Sensors}, vol.~22, no.~18, p. 6780, 2022.

\bibitem{Dey2026DataCentricDR}
S.~Dey, Z.~Khan, T.~A. PramodKumar, B.~U. Shankar, A.~K. Dhara, R.~Rajalakshmi, R.~Raman, and S.~Mitra, ``Managing diabetic retinopathy with deep learning: A data centric overview,'' \emph{arXiv preprint arXiv:2604.02448}, 2026.

\bibitem{Kwon2022DataScarcity}
G.~Kwon, E.~Kim, S.~Kim, S.~Bak, M.~Kim, and J.~Kim, ``Bag of tricks for developing diabetic retinopathy analysis framework to overcome data scarcity,'' \emph{arXiv preprint arXiv:2210.09558}, 2022.

\bibitem{Inamullah2024}
Inamullah, S.~Hassan, S.~B. Belhaouari, and I.~Amin, ``Deciphering the impact of diversity in cnn-based ensembles on overcoming data imbalance and scarcity in medical datasets: A case study on diabetic retinopathy,'' \emph{Informatics in Medicine Unlocked}, vol.~49, p. 101557, 2024.

\bibitem{djoumessi2025inherently}
K.~Djoumessi, Z.~Huang, L.~K{\"u}hlewein, A.~Rickmann, N.~Simon, L.~M. Koch, and P.~Berens, ``An inherently interpretable ai model improves screening speed and accuracy for early diabetic retinopathy,'' \emph{PLOS Digital Health}, vol.~4, no.~5, p. e0000831, 2025.

\bibitem{selvaraju2017}
R.~R. Selvaraju, M.~Cogswell, A.~Das, R.~Vedantam, D.~Parikh, and D.~Batra, ``{Grad-CAM}: Visual explanations from deep networks via gradient-based localization,'' in \emph{Proceedings of the IEEE International Conference on Computer Vision}, 2017, pp. 618--626.

\bibitem{Quellec2016DeepImageMining}
G.~Quellec, K.~Charri{\`e}re, Y.~Boudi, B.~Cochener, and M.~Lamard, ``Deep image mining for diabetic retinopathy screening,'' \emph{Medical Image Analysis}, vol.~39, pp. 178--193, 2017.

\bibitem{ghassemi2021false}
M.~Ghassemi, L.~Oakden-Rayner, and A.~L. Beam, ``The false hope of current approaches to explainable artificial intelligence in health care,'' \emph{The lancet digital health}, vol.~3, no.~11, pp. e745--e750, 2021.

\bibitem{Zhou2022AutoMorph}
Y.~Zhou, S.~K. Wagner, M.~A. Chia, A.~Zhao, P.~Woodward-Court, M.~Xu, R.~R. Struyven, D.~C. Alexander, and P.~A. Keane, ``{AutoMorph}: Automated retinal vascular morphology quantification via a deep learning pipeline,'' \emph{Translational Vision Science \& Technology}, vol.~11, no.~7, p.~12, 2022.

\bibitem{huang2021}
Y.~Huang, L.~Lin, P.~Cheng, J.~Lyu, and X.~Tang, ``Lesion-based contrastive learning for diabetic retinopathy grading from fundus images,'' in \emph{Medical Image Computing and Computer Assisted Intervention -- MICCAI 2021}.\hskip 1em plus 0.5em minus 0.4em\relax Springer, 2021, pp. 113--123.

\bibitem{cdc_nhanes}
{National Center for Health Statistics}, ``National health and nutrition examination survey: Questionnaires, datasets, and related documentation,'' \url{https://wwwn.cdc.gov/nchs/nhanes/}, 2026, accessed July 2026.

\bibitem{gargeya2017}
R.~Gargeya and T.~Leng, ``Automated identification of diabetic retinopathy using deep learning,'' \emph{Ophthalmology}, vol. 124, no.~7, pp. 962--969, 2017.

\bibitem{pearl2009causality}
J.~Pearl, \emph{Causality: Models, Reasoning, and Inference}.\hskip 1em plus 0.5em minus 0.4em\relax Cambridge University Press, 2009.

\bibitem{imai2010mediation}
K.~Imai, L.~Keele, and D.~Tingley, ``A general approach to causal mediation analysis,'' \emph{Psychological Methods}, vol.~15, no.~4, pp. 309--334, 2010.

\bibitem{hastie1990gam}
T.~J. Hastie and R.~J. Tibshirani, \emph{Generalized Additive Models}.\hskip 1em plus 0.5em minus 0.4em\relax Chapman and Hall, 1990.

\bibitem{wood2017gam}
S.~N. Wood, \emph{Generalized Additive Models: An Introduction with R}.\hskip 1em plus 0.5em minus 0.4em\relax CRC Press, 2017.

\bibitem{aptos2019-blindness-detection}
Karthik, Maggie, and S.~Dane, ``Aptos 2019 blindness detection,'' \url{https://kaggle.com/competitions/aptos2019-blindness-detection}, 2019, kaggle.

\end{thebibliography}
\end{document}